\documentclass[runningheads]{llncs}

\usepackage[utf8]{inputenc} % allow utf-8 input
\usepackage[T1]{fontenc}    % use 8-bit T1 fonts
\usepackage{hyperref}       % hyperlinks
\usepackage{url}            % simple URL typesetting
\usepackage{booktabs}       % professional-quality tables
\usepackage{multirow}
\usepackage{amsfonts,amsmath}       % blackboard math symbols
\usepackage{nicefrac}       % compact symbols for 1/2, etc.
\usepackage{microtype}      % microtypography
\usepackage{graphicx}
\usepackage[capitalise]{cleveref}
\usepackage{todonotes}
\usepackage{pifont}% http://ctan.org/pkg/pifont

\usepackage{float}
\usepackage{color}
\usepackage{tabularray}

\usepackage{hyperref}       % hyperlinks
\usepackage{sidecap}
\hypersetup{
    colorlinks,
    linkcolor={blue!50!black},
    citecolor={blue!50!black},
    urlcolor={blue!50!black}
}

\begin{document}

\title{Exploring the interplay of label bias with subgroup size and separability: A case study in mammographic density classification}

\titlerunning{Exploring the interplay of label bias with subgroup size and separability}


\author{
Emma A. M. Stanley\inst{1,2,3,4}\thanks{Work performed as a visiting student at Imperial College London.} 
\and Raghav Mehta\inst{5}
\and Mélanie Roschewitz\inst{5} 
\and Nils D. Forkert\inst{2,3,4}
\and Ben Glocker\inst{5}}

\authorrunning{E. A. M. Stanley et al.}

\institute{Department of Biomedical Engineering, University of Calgary, Canada
\and Department of Radiology, University of Calgary, Canada 
\and Hotchkiss Brain Institute, University of Calgary, Canada 
\and Alberta Children's Hospital Research Institute, University of Calgary, Canada 
\and Department of Computing, Imperial College London, UK 
\\\email{emma.stanley@ucalgary.ca}}

% \author{Anonymized Authors}  %% Added for anonymized MICCAI 2025 submission
% \authorrunning{Anonymized Author et al.}
% \institute{Anonymized Affiliations \\
%     \email{email@anonymized.com}}

\maketitle              % typeset the header of the contribution
\begin{abstract}
Systematic mislabelling affecting specific subgroups (\textit{i.e.,} label bias) in medical imaging datasets represents an understudied issue concerning the fairness of medical AI systems. In this work, we investigated how size and separability of subgroups affected by label bias influence the learned features and performance of a deep learning model. Therefore, we trained deep learning models for binary tissue density classification using the EMory BrEast imaging Dataset (EMBED), where label bias affected separable subgroups (based on imaging manufacturer) or non-separable `pseudo-subgroups'. We found that simulated subgroup label bias led to prominent shifts in the learned feature representations of the models. Importantly, these shifts within the feature space were dependent on both the relative size and the separability of the subgroup affected by label bias. We also observed notable differences in subgroup performance depending on whether a validation set with clean labels was used to define the classification threshold for the model. For instance, with label bias affecting the majority separable subgroup, the true positive rate for that subgroup fell from 0.898, when the validation set had clean labels, to 0.518, when the validation set had biased labels. Our work represents a key contribution toward understanding the consequences of label bias on subgroup fairness in medical imaging AI. 

\end{abstract}
\section{Introduction}

%expanded intro: 
An implicit assumption when training artificial intelligence (AI) models for supervised classification tasks is that the labels provided with a training dataset represent the ground truth. However, the annotations for large-scale datasets, which are required for effectively training deep models, are costly to acquire, and may not be fully reliable due to factors such as human error, subjective or ambiguous labelling tasks, or inaccurate automated labelling systems \cite{tong_xiao_learning_2015}. Incorrectly labelled data---`label noise'---has been shown to negatively impact generalizability and training dynamics \cite{frenay_classification_2014}, comprising an often overlooked but potentially pervasive problem. Medical imaging datasets may be particularly susceptible to label noise, due to inter-annotator variability \cite{yang_assessing_2023,nichyporuk_rethinking_2022}, the use of language models for automatically extracting labels from free-text radiology reports \cite{jain_visualchexbert_2021}, and the inherent complexity of many imaging-based diagnostic tasks as primary contributing factors \cite{shi_survey_2024}. 

Label noise may be present throughout a dataset, but it may also systematically affect certain subsets of data. For instance, a low-quality scanner being used in an under-resourced setting \cite{mollura_artificial_2020} may produce images that are more challenging to identify lesions from. Annotations by a less experienced radiologist at a medical center that primarily serves a particular population demographic could result in a higher rate of diagnostic errors affecting that population \cite{zhang_diagnostic_2023}.
Human decision-making biases could lead to higher rates of mis- or under-diagnosis for some genders or racial subgroups \cite{markowitz_gender_2022} within a dataset. A language model that extracts disease labels from radiology reports may perform inaccurately on reports written in different languages \cite{jones_causal_2024}. 
% can add or replace something with a concept shift example if preferred \cite{chen_algorithmic_2023}
In any of the aforementioned examples, AI models trained on such datasets could learn an inconsistent mapping between image features and labels that systematically affects particular subpopulations, possibly leading to performance disparities between subgroups within the data. 

While progress has been made toward identifying and addressing label noise in medical imaging datasets \cite{shi_survey_2024,wei_deep_2024}, little attention has been paid to the implications of systematic subgroup label noise (\textit{i.e.,} label bias) as a potential fairness issue \cite{petersen_path_2023}. In this context, our work aims to investigate the impacts of subgroup label bias on a deep learning model, using tissue density classification from mammography data as an example application of clinical importance. Specifically, we \textbf{(1)} examine the feature space of a deep learning model trained under simulated subgroup label bias, illustrating how subgroup size and separability influence learned representations, and \textbf{(2)} demonstrate notable differences in subgroup classification performance depending on whether an unbiased validation set is used for defining the classification threshold.

\section{Materials and methods}

\textbf{2.1 Dataset.} We investigated label bias in a breast tissue density classification task using full-field digital mammography images from the EMory BrEast imaging Dataset (EMBED), which were acquired by four institutional hospitals over a seven-year period  \cite{jeong2023embed}. The dataset was filtered to remove laterality mismatches, and images with spot compression and magnification. Only mediolateral oblique and craniocaudal views from patients identified as female who were assigned a single BI-RADS tissue density label of either $A$, $B$, $C$, or $D$ were used. A density label of $A$ corresponds to the lowest density (\textit{i.e.,} mostly fatty tissue), while $D$ corresponds to the highest density (\textit{i.e.,} mostly dense tissue). For the classification task, we binarized the density labels into \texttt{0} $:=\{A, B\}$ and \texttt{1} $:=\{C, D\}$, and performed undersampling to balance the two classes. The dataset was further filtered to include only the three largest imaging system manufacturers: Hologic (\texttt{HOLO}), GE Medical Systems (\texttt{GEMS}), and Fujifilm (\texttt{FUJI}). \texttt{HOLO} comprised a large majority of the dataset, followed by \texttt{GEMS} and \texttt{FUJI} (see Table \ref{table: Table 1}).

\begin{table}
\centering
\caption{Number of images belonging to each manufacturer subgroup.}
\resizebox{1\linewidth}{!}{
\begin{tblr}{
  cells = {c},
  cell{1}{1} = {r=2}{},
  cell{1}{2} = {r=2}{},
  cell{1}{3} = {r=2}{},
  cell{1}{4} = {c=4}{},
  hline{1,6} = {-}{0.08em},
  hline{2-3} = {4-7}{black},
  hline{3} = {1-3}{},
}
\textbf{Manufacturer}     & \textbf{Total} & {\textbf{Percent}\\\textbf{of Dataset}} & \textbf{Tissue Density~} &            &            &            \\
                          &                &                                         & \textbf{A}               & \textbf{B} & \textbf{C} & \textbf{D} \\
Hologic (HOLO)            & 170,995         & 89.0                                    & 18,367                    & 67,124      & 76,720      & 8,784       \\
GE Medical Systems (GEMS) & 15,965          & 8.3                                     & 636                      & 7,210       & 7,590       & 529        \\
Fujifilm (FUJI)~          & 5,130           & 2.7                                     & 425                      & 2,283       & 2,260       & 162        
\end{tblr}}
\label{table: Table 1}
\end{table}

\textbf{2.2 Subgroup label bias.} We define subgroup label bias as the \textit{systematic mislabelling of images affecting a single subgroup}. This was modelled by changing the binary tissue density labels for images in a subgroup (\textit{e.g.,} from a particular manufacturer) from class \texttt{1} to class \texttt{0}. However, we only introduced label bias to patients with tissue density category $C$, as we assumed this approach to be more realistic for potentially ambiguous cases ($B$/$C$) to be mislabelled, compared to categorical extremes ($A$/$D$). Furthermore, since many patients in the dataset had multiple images resulting from subsequent examinations, we assumed that label bias affected all images from a given patient. Therefore, in all experiments, subgroup label bias was represented as the change of binary class label \texttt{1} to \texttt{0} in 30\% of patients who were assigned a tissue density of $C$ within a given subgroup. We analyzed label bias applied to each of the three manufacturer subgroups, as well as in `pseudo-subgroups' to evaluate the role of subgroup separability \cite{jones2023separability} in label bias scenarios (see Sec. 3.1).

\textbf{2.3 Models and training.} A ResNet-18 \cite{he_deep_2016} was used to classify images into the binary tissue density classes. The dataset was randomly split into 60\%/20\%/20\% for training, validation, and testing, ensuring no patient overlap between splits. Images were preprocessed by masking out non-tissue regions and normalizing pixel values to the range [0,1]. Data augmentation for training included gamma correction, brightness/contrast jitter, and random affine transformations. Models were trained at a learning rate of $1\times10^{-5}$ with class-balanced batch sampling. The model with the highest area under the receiver operating characteristic curve (AUC) on the validation set over 10 epochs was evaluated.

\textbf{2.4 Feature inspection.} For the models trained on each label bias scenario, we performed an exploration of the learned features following the methods described in Glocker et al. \cite{glocker2023algorithmic}. Briefly described, the test set was passed through the trained model and the features from the penultimate layer were extracted. Dimensionality reduction was performed using principal component analysis (PCA). We visualized these features with kernel density estimation (KDE) plots of the first PCA mode (PC1), as it was most predictive of tissue density.

\textbf{2.5 Classification performance.} We computed subgroup-specific true (TPR) and false positive rates (FPR) for the test set \textit{using true (clean) class labels} at a threshold defined by 10\% overall FPR on the validation set. Since validation data may also be affected by label bias in a real-world setting, we compared subgroup performance between thresholds selected using both clean labels and biased labels (as described in Sec. 2.2).

\section{Results}

\textbf{3.1 Subgroup separability.}
In this work, subgroup separability was defined as the ability of a deep learning model to detect which subgroup an image belongs to \cite{jones2023separability}. To determine subgroup separability, we trained a model using the same method described in Sec. 2.3 to classify subgroups. One-vs-rest AUC for each manufacturer subgroup (\texttt{HOLO}, \texttt{GEMS}, and \texttt{FUJI}) was 1.00, demonstrating that these subgroups were fully separable. To compare to the non-separable case, we created three equally-sized `pseudo-subgroups', where subgroup labels were randomly assigned to each image in the dataset, corresponding to pseudo-subgroups \texttt{1, 2,} and \texttt{3}. A model trained to classify these subgroups achieved only chance-level AUC values of 0.50, indicating that these subgroups were not separable (\textit{i.e.,} the model was not able to distinguish the three pseudo-subgroups from one another).

\begin{figure}[H]
    \centering
    \includegraphics[width=1\linewidth]{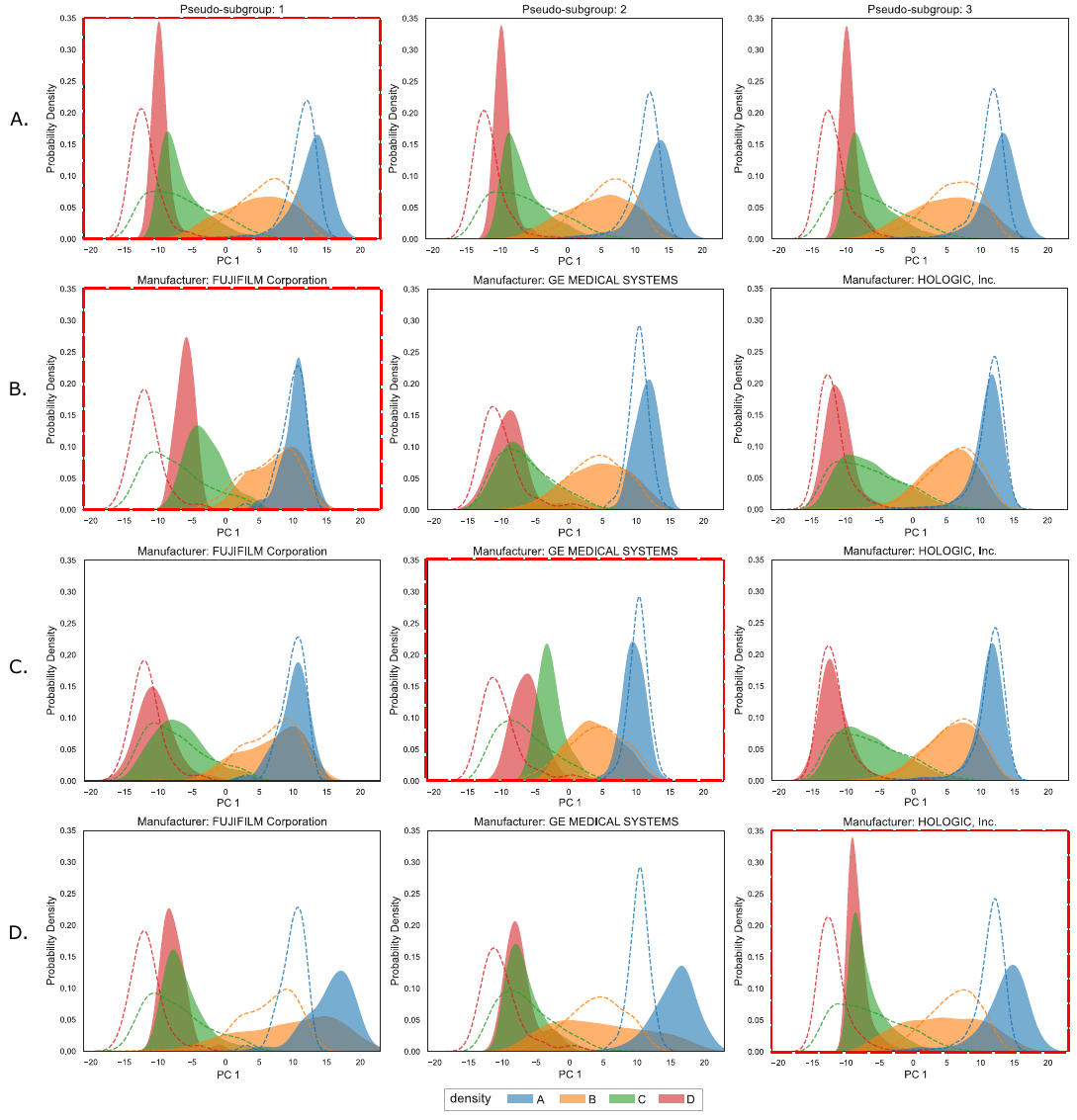}
    \caption{Distribution of tissue density categories on the first principal component (PC1) of the feature space for each label bias scenario. Row A: label bias in pseudo-subgroup 1 (\texttt{PS1}), Row B: label bias in Fujifilm (\texttt{FUJI}), Row C: label bias in GE Medical Systems (\texttt{GEMS}), Row D: label bias in Hologic (\texttt{HOLO}). Dashed outlines show the distribution of tissue density categories for the specified subgroup from the model trained with clean labels. \textcolor{red}{Red frames} indicate the subgroup with label bias.}
    \label{fig:Figure 1}
\end{figure}

\textbf{3.2 Impact of subgroup label bias on learned features.} 
Models were trained on the dataset with clean labels, representing the baseline, as well as on the datasets with label bias applied to pseudo-subgroup \texttt{1} (\texttt{PS1}) and to each manufacturer subgroup separately. KDE plots of the first principal component (PC1) of the learned features for the subgroups in each label bias scenario (\texttt{PS1}, \texttt{FUJI}, \texttt{GEMS}, and \texttt{HOLO}) are visualized in Fig. \ref{fig:Figure 1}, with the corresponding subgroup feature distribution for the clean label baseline overlaid. Notably, although the model was only trained to predict binary density classes, the feature space along PC1 in the clean label scenario naturally organized into a relatively symmetric, ordinal pattern corresponding to tissue density categories $A$, $B$, $C$, and $D$.

\textit{Non-separable subgroups.} When label bias was applied to density category $C$ in \texttt{PS1}, the features for the entire class \texttt{1} (\textit{i.e.,} density categories $C$ and $D$) shifted and clustered more closely towards class \texttt{0}. This can be seen in Fig. \ref{fig:Figure 1}A as a higher concentration of the class \texttt{1} tissue density categories near the boundary separating the two classes, as compared to the clean label baseline. In this case, where label bias was applied to a non-separable subgroup (\texttt{PS1}), this feature shift was consistent across all pseudo-subgroups (\texttt{PS1}, \texttt{PS2}, \texttt{PS3}) (cf. Fig. \ref{fig:Figure 1}A subplots).

\textit{Separable subgroups.} Figs. \ref{fig:Figure 1}B and \ref{fig:Figure 1}C illustrate the impact of label bias applied to \texttt{FUJI} and \texttt{GEMS} respectively, both fully separable but minority subgroups. Here, it can be seen that the feature shift occurred primarily in the subgroup with label bias. The feature space of \texttt{HOLO} remained similar to the clean label baseline, and the feature space of the other minority subgroup experienced small perturbations but did not undergo a feature shift as prominent as the subgroup with label bias. 

In contrast, when label bias was applied to \texttt{HOLO}, a separable subgroup that comprised a large majority of the dataset, the feature shift along PC1 of the class with label bias was evident across all subgroups (cf. Fig. \ref{fig:Figure 1}D). In particular, while \texttt{HOLO} had the highest concentration of class \texttt{1} features near to class \texttt{0} (represented by the sharpest peak), \texttt{GEMS} and \texttt{FUJI} also had features for both $C$ and $D$ categories shift toward class \texttt{0}. In all subgroups, the two class \texttt{1} density categories overlapped, as opposed to the ordinal pattern from the clean label baseline. It can also be seen that the distribution of class \texttt{0} tissue density categories had a wider spread, compared to the other label bias scenarios.

\textbf{3.3 Impact of subgroup label bias on performance.} TPR and FPR for the three manufacturer subgroups in each label bias scenario are presented in Table \ref{table: Table 2}. Importantly, using a validation set with clean labels to define an operating point threshold led to different subgroup performance values compared to using a validation set with biased labels, depending on separability and size of the subgroup with label bias.

\begin{table}[H]
\centering
\caption{True positive rate (TPR) and false positive rate (FPR) for each subgroup under each label bias scenario. Thresholds were computed at 10\% overall FPR on the validation set, using either clean or biased labels. Values in parentheses indicate percent change relative to the clean baseline.}
\resizebox{\linewidth}{!}{%
\begin{tblr}{
  cells = {c},
  column{5} = {rightsep=25pt},
  cell{1}{1} = {r=3}{},
  cell{1}{2} = {r=3}{},
  cell{1}{3} = {c=3}{},
  cell{1}{6} = {c=3}{},
  cell{2}{3} = {c=6}{},
  cell{4}{1} = {c=2}{},
  cell{7}{1} = {r=2}{},
  cell{7}{2} = {r=2}{},
  cell{7}{3} = {c=6}{},
  cell{9}{1} = {c=2}{},
  hline{1,16} = {-}{0.08em},
  hline{2-4,8-9} = {3-8}{black},
  hline{4,9} = {1-2}{},
  hline{5,10} = {-}{black},
  hline{6,13} = {-}{dotted},
  hline{7} = {-}{},
}
{\textbf{Subgroup with }\\\textbf{Label Bias}} & {\textbf{Validation}\\\textbf{Set Labels}} & \textbf{TPR}                     &                 &                 & \textbf{FPR}    &                 &                  \\
                                               &                                             & \textbf{Non-Separable Subgroups} &                 &                 &                 &                 &                  \\
                                               &                                             & \textbf{PS1}                     & \textbf{PS2}    & \textbf{PS3}    & \textbf{PS1}    & \textbf{PS2}    & \textbf{PS3}     \\
Clean (Baseline)                               &                                             & 0.916                            & 0.913           & 0.915           & 0.106           & 0.104           & 0.104            \\
PS1                                            & Clean                                       & 0.910 \tiny{(-00.7\%)}                  & 0.912 \tiny{(-00.1\%)} & 0.912 \tiny{(-00.3\%)}  & 0.106 \tiny{(+00.0\%)} & 0.108 \tiny{(+03.8\%)} & 0.104 \tiny{(+00.0\%)}  \\
PS1                                            & Biased                                      & 0.826 \tiny{(-09.8\%)}                  & 0.828 \tiny{(-09.3\%)} & 0.826 \tiny{(-08.6\%)}  & 0.042 \tiny{(-60.4\%)} & 0.045 \tiny{(-56.7\%)} & 0.048 \tiny{(-53.8\%)}  \\
~                                              & ~                                           & \textbf{Separable Subgroups}     &                 &                 &                 &                 &                  \\
                                               &                                             & \textbf{FUJI}                    & \textbf{GEMS}   & \textbf{HOLO}   & \textbf{FUJI}   & \textbf{GEMS}   & \textbf{HOLO}    \\
Clean (Baseline)                               &                                             & 0.923                            & 0.914           & 0.915           & 0.069           & 0.170           & 0.100            \\
FUJI                                           & Clean                                       & 0.750 \tiny{(-18.7\%)}                  & 0.907 \tiny{(-00.8\%)} & 0.914 \tiny{(-00.1\%)} & 0.013 \tiny{(-81.2\%)} & 0.158 \tiny{(-07.1\%)} & 0.106 \tiny{(+06.0\%)}  \\
GEMS                                           & Clean                                       & 0.941 \tiny{(+02.0\%)}                  & 0.780 \tiny{(-14.7\%)} & 0.914 \tiny{(-00.1\%)} & 0.092 \tiny{(+33.3\%)} & 0.087 \tiny{(-48.8\%)} & 0.103 \tiny{(+03.0\%)}  \\
HOLO                                           & Clean                                       & 0.938 \tiny{(+01.6\%)}                  & 0.943 \tiny{(-03.2\%)} & 0.898 \tiny{(-01.9\%)} & 0.069 \tiny{(+00.0\%)} & 0.228 \tiny{(+34.1\%)} & 0.096~ \tiny{(-04.0\%)} \\
FUJI                                           & Biased                                      & 0.728 \tiny{(-21.1\%)}                  & 0.904 \tiny{(-01.1\%)} & 0.910 \tiny{(-00.5\%)} & 0.013 \tiny{(-81.2\%)} & 0.153 \tiny{(-10.0\%)} & 0.102 \tiny{(+02.0\%)}  \\
GEMS                                           & Biased                                      & 0.931 \tiny{(+00.9\%)}                  & 0.726 \tiny{(-20.6\%)} & 0.901 \tiny{(-01.5\%)} & 0.076 \tiny{(+10.1\%)} & 0.068 \tiny{(-60.0\%)} & 0.089 \tiny{(-11.0\%)}  \\
HOLO                                           & Biased                                      & 0.844 \tiny{(-08.6\%)}                  & 0.828 \tiny{(-09.4\%)} & 0.518 \tiny{(-43.4\%)} & 0.015 \tiny{(-78.3\%)} & 0.080 \tiny{(-52.9\%)} & 0.012 \tiny{(-88.0\%)}  
\end{tblr}}
\label{table: Table 2}
\end{table}

\textit{Non-separable subgroups.} When clean labels were used for computing the threshold with label bias applied to \texttt{PS1} during training, subgroup TPR and FPR for all subgroups were very similar to the clean label baseline. However, when biased labels were used for computing this threshold, TPR for all subgroups decreased by nearly 0.10, and FPR decreased by between 0.05-0.07.

\textit{Separable subgroups.} When label bias affected minority separable subgroups (\textit{i.e.,} \texttt{FUJI}  or \texttt{GEMS}), the TPR/FPR decreased for the affected subgroup, regardless of whether the threshold was set on a clean or biased-label validation set. For the subgroups without label bias, performance stayed relatively close to baseline, though usually slighly lower when the threshold was computed using the biased-label validation set. In contrast, the largest difference in subgroup performance between clean and biased validation sets resulted from label bias affecting \texttt{HOLO}, the majority separable subgroup. With a clean label validation set, TPR and FPR for \texttt{HOLO} decreased only slightly compared to the baseline. In contrast, when the classification threshold was computed using biased labels in the validation set, TPR for \texttt{HOLO} fell to 0.518, and TPR for \texttt{GEMS} and \texttt{FUJI} were also reduced by around 0.10 compared to when clean labels were used. FPR for all separable subgroups also decreased substantially.

\section{Discussion}
We demonstrated that the impact of subgroup label bias in a deep learning model trained for a binary classification task depended on both separability and relative size of the affected subgroup. In general, label bias led to a feature shift along PC1, in which features for the class affected by label bias shifted toward the other class. When label bias affected a non-separable subgroup, this feature shift occurred across all other non-separable subgroups. In contrast, when label bias was applied to a separable minority subgroup, this feature shift was primarily evident within the affected subgroup, possibly because the model learned that this `noisy' mapping of images to labels was only associated with features of that particular subgroup. However, with label bias applied to the majority separable subgroup, all subgroups experienced a feature shift, similar to the non-separable case. This may be due to the much higher frequency that noisy labels were observed during training in this case, effectively causing the model to associate this noisy mapping with the entire dataset. 

\begin{figure}[H]
    \centering
    \includegraphics[width=1\linewidth]{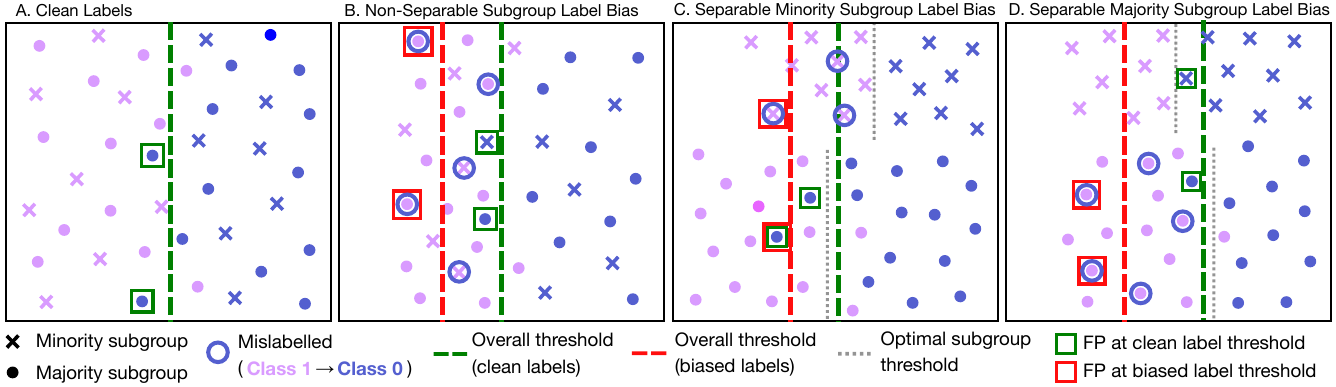}
    \caption{Toy example illustrating how feature shifts caused by label bias could impact subgroup performance, when the operating point threshold of the model is selected using overall FPR on the validation set.}
    \label{fig:Figure 2}
\end{figure}

We speculate that the feature shifts observed in PC1 likely also led to a shift of optimal class separation thresholds in the high-dimensional decision space of the model. Fig. \ref{fig:Figure 2} contains a simple toy example that illustrates how feature and threshold shifts caused by label bias could help to explain the impacts on subgroup performance presented in Table \ref{table: Table 2}.

For example, consider label bias affecting a non-separable subgroup (Fig. \ref{fig:Figure 2}B). In this case, selecting an operating point threshold using the clean labels would still lead to high classification performance, even though the feature space (and the corresponding optimal class decision threshold) would have shifted. However, using a validation set with biased labels would move the selected operating point threshold toward class \texttt{1}, since calculation of the overall FPR would take mislabelled images into account. This would result in lower TPR and FPR for all subgroups, similar to what we observed when label bias affected \texttt{PS1}.

With label bias in a minority separable subgroup (Fig. \ref{fig:Figure 2}C), we observed a feature shift predominantly in the affected subgroup. However, since the classification threshold is based on overall FPR, this threshold would be primarily determined by the majority subgroup. Therefore, with a clean validation set, TPR and FPR for the minority subgroup would decrease since the optimal threshold for that subgroup would be shifted toward class \texttt{0}, compared to the overall threshold. Using biased labels for validation, the overall threshold would be shifted slightly more toward class \texttt{1}, leading to further decreased TPR and FPR for all subgroups, as we observed when label bias affected \texttt{GEMS} and \texttt{FUJI}.

In the case of label bias affecting the majority subgroup (Fig \ref{fig:Figure 2}D), the classification threshold would be shifted toward class \texttt{0}, but performance for the majority group would remain similar to the baseline. Assuming the optimal threshold for the majority subgroup shifted more than for the minority subgroup, TPR and FPR for the minority groups would increase. On the other hand, using biased labels for the overall FPR calculation would move the operating point threshold much further in the class \texttt{1} direction to account for the mislabelled images in the majority subgroup. This would then lead to a drastic decrease in TPR and FPR, particularly for the majority subgroup, as we observed when label bias was applied to \texttt{HOLO}.

\section{Conclusions}
The impact of subgroup label bias in medical imaging datasets is under-explored, but has important implications in the context of AI fairness. It has been shown that sociodemographic subgroups in medical imaging datasets can have a range of separability, for example, sex in retinal fundus images may have relatively low separability \cite{jones2023separability}, while self-reported race subgroups in chest x-rays can be nearly completely separable \cite{gichoya2022ai}. Furthermore, substantial differences in relative subgroup size are often present, particularly within self-reported race categories \cite{chen_algorithmic_2023}. Our results indicate that subgroup separability and size have distinct impacts on group fairness under label bias and are, therefore, both crucial to consider in future investigations. From a clinical usability standpoint, where it is necessary to define classification thresholds, our work emphasizes the importance of understanding whether datasets used for validating models are affected by label bias. In particular, defining an operating point using biased labels could substantially impact performance for one or all subgroups, if the affected subgroup is a large majority or if subgroups are non-separable. Therefore, future work should not only further evaluate the impacts of label bias in controlled scenarios, but also continue to work toward developing methods for reliably detecting the presence of label bias in medical imaging datasets. 

\textit{Limitations.} It is important to note that we assumed that the tissue density labels in the dataset were initially clean (\textit{i.e.,} that there was no significant existing label bias). Furthermore, it is possible that other separable subgroups existed in the dataset, which may have introduced interacting effects. We explored only the first principal component of the feature space, therefore, confirming that the trends illustrated in the toy example correspond to the high-dimensional decision space of the model is a next step in future work. We studied separable subgroups defined by hardware-related differences, and only considered a static amount (30\%) of label bias in one direction (\textit{e.g.,} class \texttt{1} to class \texttt{0}). Future work should investigate varying degrees of label bias affecting both classes, and whether the observed trends hold for label bias in sociodemographic subgroups.

\textit{Reproducibility.} Code for data processing, experiments, and analysis is available at \href{https://github.com/biomedia-mira/mammo-label-bias}{https://github.com/biomedia-mira/mammo-label-bias}.

\subsubsection{Acknowledgments.} 
E.A.M.S. and N.D.F. acknowledge support from the Natural Science and Engineering Research Council of Canada, the Killam Trusts, Alberta Innovates, the Canada Research Chairs Program, and the River Fund at Calgary Foundation. M.R. is funded by an Imperial College London President’s PhD Scholarship and a Google PhD Fellowship. R.M. is funded by the European Union’s Horizon Europe research and innovation programme under grant agreement 101080302. B.G. received support from the Royal Academy of Engineering as part of his Kheiron/RAEng Research Chair.

\subsubsection{Disclosure of Interests.} 
B.G. is part-time employee of DeepHealth. No other competing interests.

\bibliographystyle{splncs04}
\bibliography{references}

\end{document}